\documentclass[conference]{IEEEtran}
\usepackage{times}

\usepackage[numbers]{natbib}
\usepackage{multicol}
\usepackage[bookmarks=true]{hyperref}
\usepackage{todonotes}
\usepackage{amsfonts}  
\usepackage{amsmath}   
\usepackage{amsthm}
\usepackage{algorithm}
\usepackage{algpseudocode}
\usepackage{booktabs}
\usepackage{multirow}
\usepackage{subfigure}

\begin{document}

\title{Linear Memory SE(2) Invariant Attention}


%
\author{\authorblockN{Ethan Pronovost,
Neha Boloor,
Peter Schleede, 
Noureldin Hendy, 
Andres Morales, and
Nicholas Roy }
\authorblockA{Zoox \\
\texttt{\{epronovost,nboloor,pschleede,nhendy,andres,nroy\}@zoox.com}}
}

\maketitle

\begin{abstract}
Processing spatial data is a key component in many learning tasks for autonomous driving such as motion forecasting, multi-agent simulation, and planning.
Prior works have demonstrated the value in using SE(2) invariant network architectures that consider only the relative poses between objects (e.g. other agents,  scene features such as traffic lanes).
However, these methods compute the relative poses for all pairs of objects explicitly, requiring quadratic memory.
In this work, we propose a mechanism for SE(2) invariant scaled dot-product attention that requires linear memory relative to the number of objects in the scene.
Our SE(2) invariant transformer architecture enjoys the same scaling properties that have benefited large language models in recent years.
We demonstrate experimentally that our approach is practical to implement and improves performance compared to comparable non-invariant architectures.
\end{abstract}

\IEEEpeerreviewmaketitle

\section{Introduction}
Learning-based methods for planning and control in the presence of dynamic agents requires modelling the spatial geometry of each agent, such as the two dimensional position and heading of each agent in SE(2).
The SE(2) geometry of each agent, along with any other agent data, can be used to learn multi-agent models for tasks such as motion forecasting \cite{gorela_cui_2023, wayformer_nayakanti_2023, query_centric_zhou_2023}, planning \cite{parting_with_misconceptions_dauner_2023, waymax_gulino_2023, scaling_all_you_need_harmel_2024}, agent simulation \cite{smart_wu_2024, realistic_agents_closed_loop_zhang_2023, closed_loop_fine_tuning_zhang_2025, behavior_gpt_zhou_2024} and scenario generation \cite{scene_diffuser_jiang_2024, scenario_diffusion_pronovost_2023, lctg_tan_2023}.
Critically, only the relative geometry between elements is needed for certain kinds of tasks such as multi-agent motion forecasting, where model architectures that are explicitly SE(2) invariant have achieved state-of-the-art performance \cite{gorela_cui_2023, query_centric_zhou_2023}.

\begin{figure}
    \centering
    \subfigure[Absolute Position: No Invariance]{
      \includegraphics[width=0.41\textwidth]{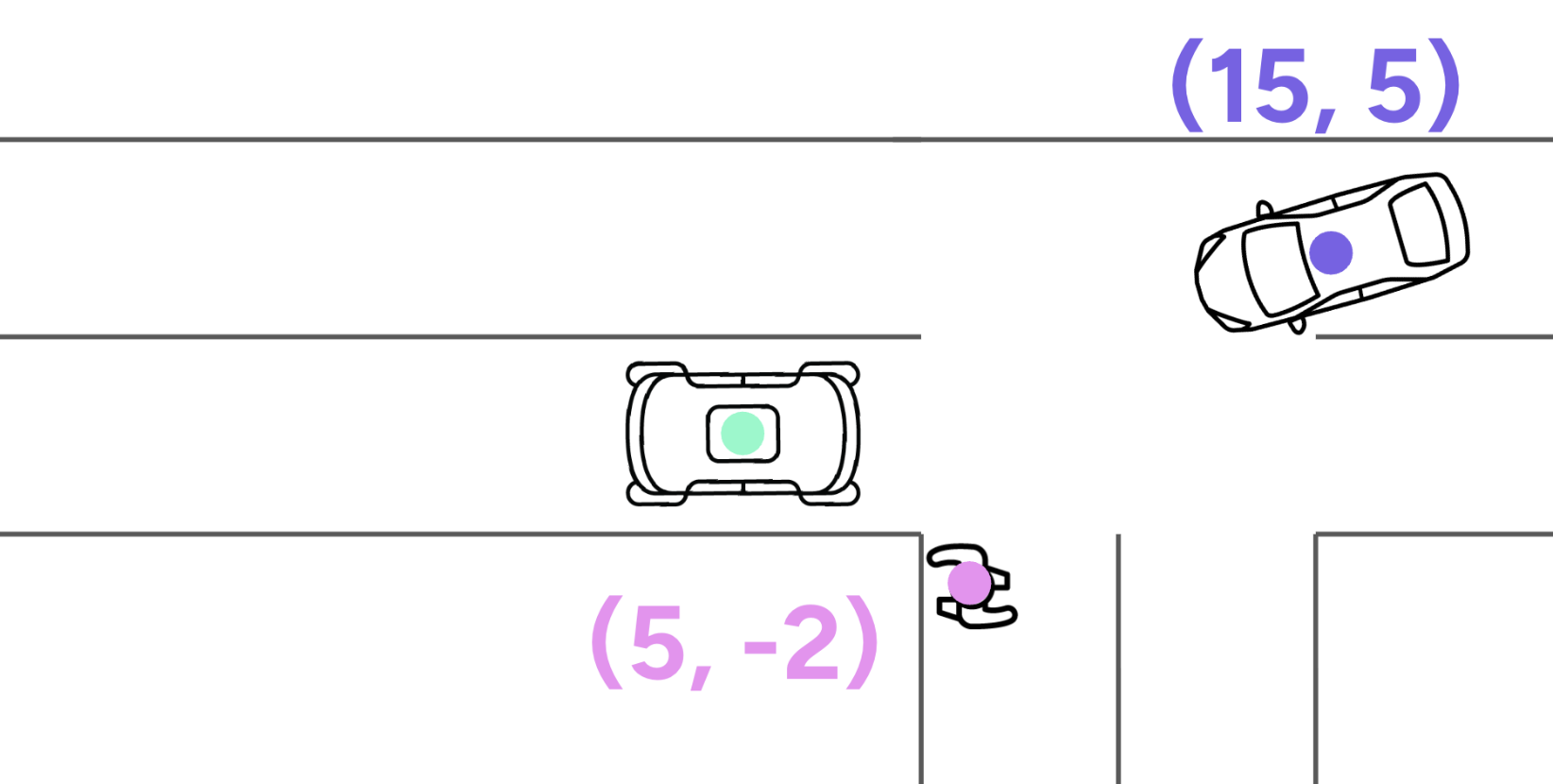}
      \label{fig:diagram-abs}
    }
    \subfigure[Relative Position: Translation Invariance]{
      \includegraphics[width=0.41\textwidth]{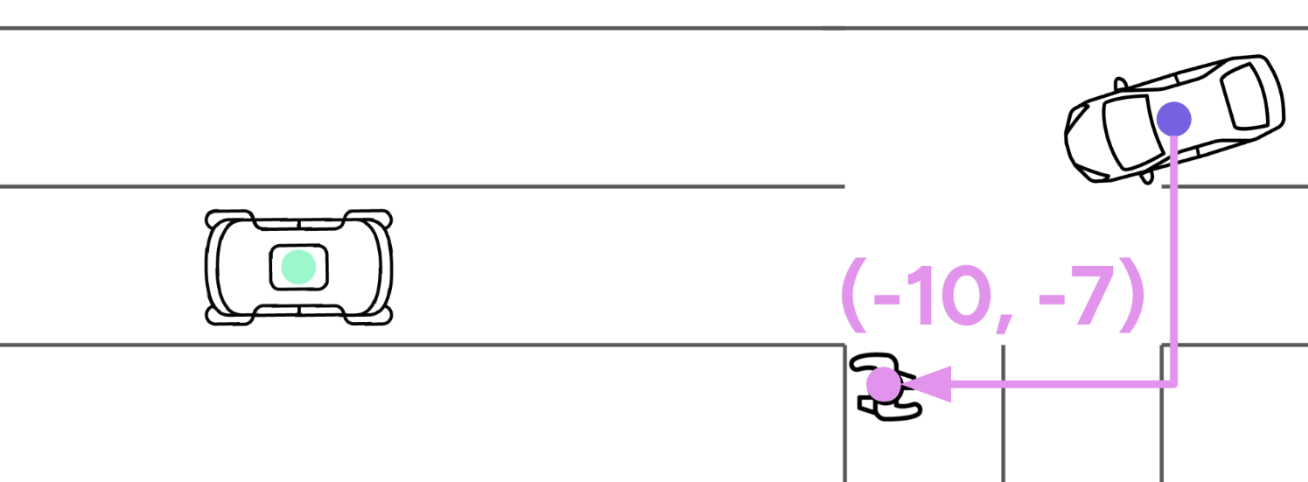}
      \label{fig:diagram-2d}
    }
    \subfigure[Relative Pose: Translation and Rotation Invariance]{
      \includegraphics[width=0.33\textwidth]{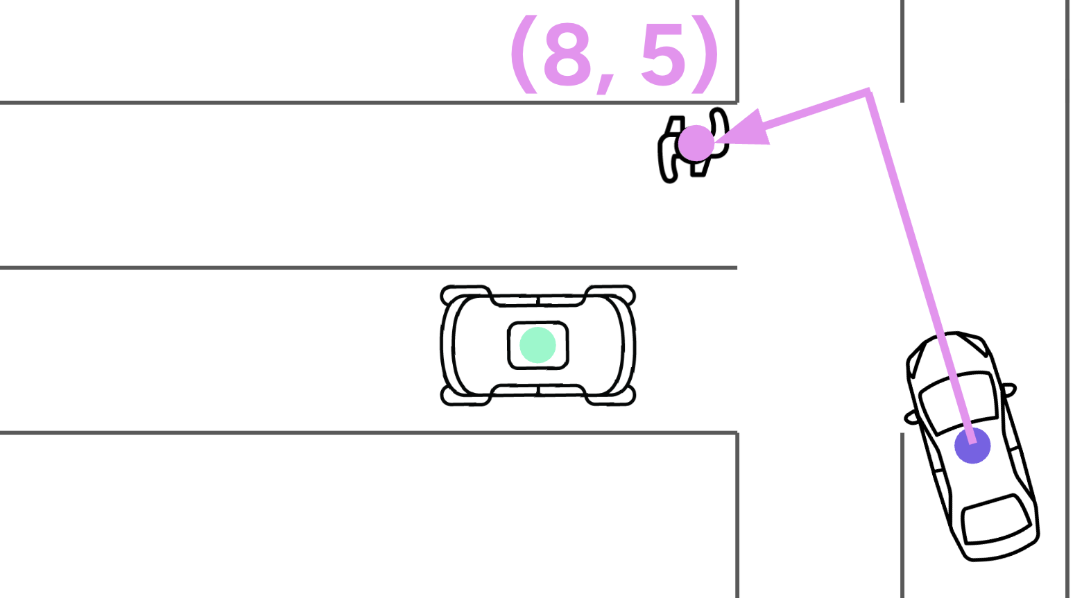}
      \label{fig:diagram-se2}
    }
    \caption{Three levels of invariance for SE(2) poses. In this example, the coordinate system is centered on the robot (green), as is standard in autonomous driving. The relationship between the car (purple) and the pedestrian (pink) should be invariant to the robot location. (a) Absolute position encodings easily require linear memory but are not invariant. (b) Relative position encodings can be implemented in linear memory with RoPE and are translation invariant, but not rotation invariant. (c) In this work we present a novel linear memory relative pose encoding that is both translation and rotation invariant.}
    \label{fig:invariance-diagram}
\end{figure}

Existing SE(2) invariant model architectures explicitly compute the relative pose between every pair of elements \cite{gorela_cui_2023, query_centric_zhou_2023}, requiring an amount of GPU high bandwidth memory (HBM) that is quadratic in the number of scene elements.
While transformer models \cite{attention_all_you_need_vaswani_2017} originally required quadratic memory as well, recent work has developed a technique for computing scaled dot-product attention (SDPA) using only linear HBM \cite{flashattention_dao_2022}.
This leads to the natural question: can we develop an SE(2) invariant transformer architecture that requires linear HBM?

Rotary position embeddings (RoPE) \cite{roformer_su_2024} provide a linear-memory attention mechanism that encodes relative positions in $\mathbb R$.
This approach has been extended to $\mathbb R^d$ and other abelian groups \cite{rope_vision_heo_2024, string_schenck_2025}.
For non-abelian groups, GTA \cite{gta_miyato_2024} proposes a related method for SE(3) and SO(2).
As we will demonstrate below, using this approach for SE(2) is not optimal for multi-agent behavior simulation.
The attention mechanism used in GTA encodes the position coordinates directly, in contrast to the sine-cosine encoding used in RoPE \cite{fourier_features_tancik_2020}.

In this work, we propose a new SE(2) invariant attention mechanism that requires linear HBM and performs better than both two-dimensional RoPE \cite{rope_vision_heo_2024} and GTA \cite{gta_miyato_2024} for multi-agent behavior simulation.
Our mechanism is equivalent to having a block-diagonal matrix of 2D rotation matrices between the query and key vectors, where the angles of those rotation matrices are based on either the relative x-position, the relative y-position, or the relative heading between the query and key.
To achieve this linear-memory property for agent behaviour,  we introduce a Fourier series approximation, and show that the approximation error can be set to less than $10^{-3}$ while still achieving reasonable inference speed.
We show that using this attention mechanism as a drop-in replacement for the standard attention mechanism in a transformer-based agent simulation model improves performance, both in an absolute sense and relative to total training time.

\begin{algorithm}[t]
\caption{Relative Scaled Dot-Product Attention with Quadratic Memory} \label{alg:relative-attention-quadratic}
\begin{algorithmic}[1]
\Require Queries and locations $\{ (\mathbf q_n, \mathbf p_n) \}_{n=1}^N$
\Require Keys, values, and locations $\{ ( \mathbf k_m, \mathbf v_m, \mathbf p_m) \}_{m=1}^M$
\Require Function $\phi : \mathcal G \to \mathbb R^{d \times d}$
\State $b_{nm} \leftarrow \mathbf q_n^\top \phi(\mathbf p_{n \to m}) \mathbf k_m \quad \forall n, m$
\State $a_{nm} \leftarrow \left. \exp \left( \frac{b_{nm}}{\sqrt{d}} \right) \middle/ \sum_{j=1}^M \exp \left( \frac{b_{nj}}{\sqrt{d}} \right) \right. \quad \forall n, m$
\State $\mathbf o_n \leftarrow \sum_{m=1}^M a_{nm} \phi (\mathbf p_{n \to m} ) \mathbf v_m \quad \forall n$
\State \Return $\{ \mathbf o_n \}_{n=1}^N$
\end{algorithmic}
\end{algorithm}

\begin{algorithm}[t]
\caption{Relative Scaled Dot-Product Attention with Linear Memory} \label{alg:relative-attention}
\begin{algorithmic}[1]
\Require Queries and locations $\{ (\mathbf q_n, \mathbf p_n) \}_{n=1}^N$
\Require Keys, values, and locations $\{ ( \mathbf k_m, \mathbf v_m, \mathbf p_m) \}_{m=1}^M$
\Require Functions $\phi_q$ and $\phi_k$ satisfying Equation \ref{eqn:phi-factorization}
\Require Standard scaled dot-product attention $\textrm{SDPA}$
\State $\widetilde {\mathbf q_n} \leftarrow \sqrt[4]{\frac{c}{d}} \cdot \phi_q (\mathbf p_n)^\top \mathbf q_n \quad \forall n$
\State $\widetilde {\mathbf k_m}, \widetilde{\mathbf v_m} \leftarrow \sqrt[4]{\frac{c}{d}} \cdot \phi_k (\mathbf p_m) \mathbf k_m, \phi_k (\mathbf p_m) \mathbf v_m \quad \forall m$
\State $\{ \widetilde {\mathbf o_n} \}_{n=1}^N = \textrm{SDPA} \left( \{ \widetilde {\mathbf q_n} \}_{n=1}^N, \{ \widetilde {\mathbf k_m} \}_{m=1}^M, \{ \widetilde {\mathbf v_m} \}_{m=1}^M \right)$
\State $\mathbf o_n \leftarrow \phi_q (\mathbf p_n) \widetilde {\mathbf o_n} \quad \forall n$
\State \Return $\{ \mathbf o_n \}_{n=1}^N$
\end{algorithmic}
\end{algorithm}

\section{Problem Formulation}

\subsection{Relative Attention}

We begin by generalizing the approach of \citet{roformer_su_2024} to other groups.
Consider a sequence of query vectors $\{ \mathbf q_n \}_{n=1}^N$, key vectors $\{ \mathbf k_m \}_{m=1}^M$, and value vectors $\{ \mathbf v_m \}_{m=1}^M$, all in $\mathbb R^d$. 
Let $\mathcal G$ be a group (in this work we consider $\mathbb R$, $\mathbb R^2$, and $\textrm{SE}(2)$).
Each query vector $\mathbf q_n$ is associated with a location $\mathbf p_n \in \mathcal G$ and likewise for $\mathbf k_m, \mathbf v_m$ and $\mathbf p_m \in \mathcal G$.
The relative location from $\mathbf p_n$ to $\mathbf p_m$ is
$\mathbf p_{n \to m} = \mathbf p_n^{-1} \mathbf p_m$.
To generalize the approach of rotary position embeddings, we define a triple of functions $\phi : \mathcal G \to \mathbb R^{d \times d}$, $\phi_q : \mathcal G \to \mathbb R^{d \times c}$, $\phi_k : \mathcal G \to \mathbb R^{c \times d}$ that satisfy
\begin{equation} \label{eqn:phi-factorization}
\phi \left( \mathbf p_{n \to m} \right) = \phi_q \left( \mathbf p_n \right) \phi_k \left( \mathbf p_m \right)
\end{equation}
for all $\mathbf p_n, \mathbf p_m \in \mathcal G$.
We can incorporate $\phi(\mathbf p_{n \to m})$ into the standard scaled dot-product attention algorithm to encode relative information, as shown in Algorithm \ref{alg:relative-attention-quadratic}.
This algorithm is invariant to transformations of $\mathcal G$: for any $\mathbf z \in \mathcal G$, if we replace all $\mathbf p_n$ with $\mathbf z^{-1} \mathbf p_n$ and all $\mathbf p_m$ with $\mathbf z^{-1} \mathbf p_m$, the output of the algorithm will not change.
\begin{equation}
\begin{split}
\forall &\mathbf z \in \mathcal G, \\
& \mathcal A_1 \left(\{ (\mathbf q_n, \mathbf p_n) \}_{n=1}^N, \{ ( \mathbf k_m, \mathbf v_m, \mathbf p_m) \}_{m=1}^M \right) = \\
& \mathcal A_1 \left (\{ (\mathbf q_n, \mathbf z^{-1} \mathbf p_n) \}_{n=1}^N, \{ ( \mathbf k_m, \mathbf v_m, \mathbf z^{-1} \mathbf p_m) \}_{m=1}^M \right)
\end{split}
\end{equation}

In practice, $\phi$ (and therefore $\phi_q$ and $\phi_k$) are chosen to be block-diagonal so each block of features from the query and key vectors is modified independently.
For simplicity we will discuss a single block at a time, with the understanding that a full implementation will stack many such blocks together to achieve larger feature dimensions.
We refer the reader to \citet{roformer_su_2024} Section 3.2.2 for an illustration of this stacked implementation.

\begin{figure}[t]
    \centering
    \includegraphics[width=0.48\textwidth]{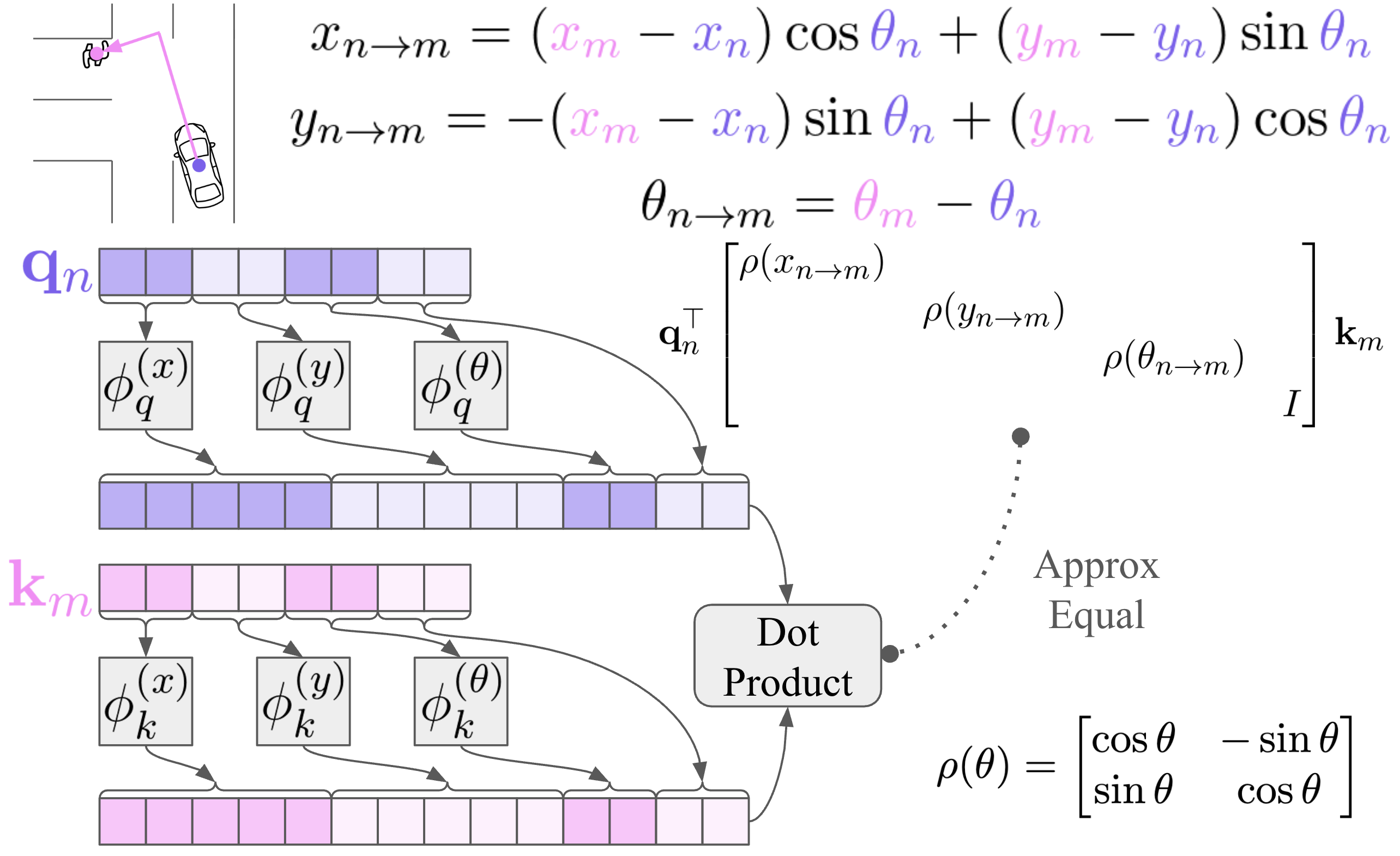}
    \caption{Overview of the approach. For each query token $\mathbf q_n$ we apply a block-diagonal linear transformation $\phi_q (\mathbf p_n)$ that depends on that token's associated pose $\mathbf p_n$, and similarly for each key token $\mathbf k_m$ with block-diagonal linear transformation $\phi_k (\mathbf p_m)$. These projected tokens $\widetilde{\mathbf q_n} = \phi_q(\mathbf p_n)^\top \mathbf q_n$ and $\widetilde{\mathbf k_m} = \phi_k (\mathbf p_m) \mathbf k_m$ are passed to a standard scaled dot-product attention kernel such as Flash Attention. By the construction of $\phi_q$ and $\phi_k$, the dot product $\widetilde{\mathbf q_n}^\top \widetilde{ \mathbf k_m}$ approximates the quadratic form $\mathbf q_n^\top M \mathbf k_m$, where $M$ is a block-diagonal matrix whose blocks are 2D rotation matrices based on the SE(2) relative x, y, and yaw between $\mathbf p_n$ and $\mathbf p_m$.}
    \label{fig:methodology}
\end{figure}

\subsection{Linear Memory Relative Attention}
Naively implementing Algorithm \ref{alg:relative-attention-quadratic} requires $O(NM)$ memory to compute $\phi(\mathbf p_{n \to m})$ for all query-key pairs.
For standard SDPA, specialized GPU kernels such as Flash Attention \cite{flashattention_dao_2022} have been developed to avoid materializing quadratic tensors.
While we could implement a specialized GPU kernel for Algorithm \ref{alg:relative-attention-quadratic}, doing so would be challenging both to write and maintain.
Instead, we can use Equation \ref{eqn:phi-factorization} to rewrite the equations of Algorithm \ref{alg:relative-attention-quadratic} to do linear memory pre-processing, standard SDPA, and linear memory post-processing.
We can then use Flash Attention without modification as a subroutine to perform the entire computation without having to materialize any $N \times M$ tensors.

We  rewrite the bilinear form in line 1 of Algorithm \ref{alg:relative-attention-quadratic} as:
\begin{equation}
\mathbf q_n^\top \phi \left( \mathbf p_{n \to m} \right) \mathbf k_m = \left( \phi_q (\mathbf p_n)^\top \mathbf q_n \right)^\top \left( \phi_k (\mathbf p_m) \mathbf k_m \right)
\end{equation}
Similarly, we rewrite the aggregation in line 3 as:
\begin{equation}
\sum_{m=1}^M a_{nm} \phi \left( \mathbf p_{n \to m} \right) \mathbf v_m  = \phi_q \left(\mathbf p_n \right) \sum_{m=1}^M a_{nm} \phi_k (\mathbf p_m) \mathbf v_m
\end{equation}
These two results are combined in Algorithm \ref{alg:relative-attention}, which matches Algorithm \ref{alg:relative-attention-quadratic} while only requiring linear HBM.
The quadratic computation happens implicitly inside the SDPA function.

This attention algorithm can replace standard SDPA in a transformer \cite{attention_all_you_need_vaswani_2017}.
The performance of such a model will heavily depend on the choice of $\phi, \phi_q, \phi_k$.
Functions that always output the zeros matrix trivially satisfy Equation \ref{eqn:phi-factorization} but would be useless for a neural network.
The following sections discuss potential choices for these functions.

\subsection{One-Dimensional: Rotary Position Embeddings}
We first describe \citet{roformer_su_2024} within this framework.
RoPE uses $\mathcal G = \mathbb R$ and 2D rotation matrices 
\begin{equation}
\rho (\theta) = \begin{bmatrix} \cos \theta & - \sin \theta \\ \sin \theta & \cos \theta \end{bmatrix}
\end{equation}
as the matrix functions: 
\begin{equation}
\begin{split}
\phi(\mathbf p_{n \to m}) &= \rho( \alpha \cdot \mathbf p_{n \to m}) \in \mathbb R^{2 \times 2} \\
\phi_q(\mathbf p_n) &= \rho(-\alpha \cdot \mathbf p_n) \in \mathbb R^{2 \times 2} \\
\phi_k (\mathbf p_m) &= \rho (\alpha \cdot \mathbf p_m) \in \mathbb R^{2 \times 2}
\end{split}
\end{equation}
for some scale factor $\alpha \in \mathbb R$.
The full implementation stacks many copies of this in a large block diagonal matrix, where each block uses a different scale factor.
This choice of rotation matrices has been found to perform well for 1-dimensional sequences \cite{roformer_su_2024, llama_touvron_2023}.
Using different scale factors allow the model to reason about different resolution features \cite{fourier_features_tancik_2020}.

\subsection{Two-Dimensional} \label{sec:2d-rope}
RoPE can be easily extended to $\mathbb R^2$ (or any other higher dimension).
Letting $\mathbf p_n = (x_n, y_n)$ and $\mathbf p_m = (x_m, y_m)$, we can combine two copies of RoPE in block-diagonal fashion: 
\begin{equation}
\begin{split}
\phi(\mathbf p_{n\to m}) &= \textrm{diag} \left[ \rho(\alpha x_{n \to m}), \rho(\alpha y_{n \to m}) \right] \in \mathbb R^{4 \times 4} \\
\phi_q(\mathbf p_n) &= \textrm{diag} \left[ \rho(-\alpha x_n), \rho(-\alpha y_n) \right]  \in \mathbb R^{4 \times 4}\\
\phi_k(\mathbf p_m) &= \textrm{diag} \left[ \rho(\alpha x_m), \rho(\alpha y_m) \right] \in \mathbb R^{4 \times 4} \\
\end{split}    
\end{equation}
This approach has been used \cite{rope_vision_heo_2024, string_schenck_2025} for vision transformers \cite{vit_dosovitskiy_2021}, and can be extended to other abelian groups \cite{string_schenck_2025}.
Applying this approach to SE(2) yields invariance to translations but not to rotations, as shown in Figure \ref{fig:diagram-2d}.

\subsection{SE(2) Representation} \label{sec:se2-rep}
Let $\mathbf p_n = (x_n, y_n, \theta_n) \in \textrm{SE}(2)$ and likewise for $\mathbf p_m$.
The most immediate option for defining $\phi$ is to use the SE(2) group representation $\psi : \textrm{SE}(2) \to \mathbb R^{3 \times 3}$ defined as
\begin{equation}
\psi(x, y, \theta) = \begin{bmatrix}
\cos \theta & - \sin \theta & x \\
\sin \theta & \cos \theta & y \\
0 & 0 & 1
\end{bmatrix}
\end{equation}
Then
\begin{equation}
\begin{split}
\phi(\mathbf p_{n \to m}) &= \psi(\mathbf p_n)^{-1} \psi(\mathbf p_m) \in \mathbb R^{3 \times 3} \\
\phi_q(\mathbf p_n) &= \psi(\mathbf p_n^{-1}) \in \mathbb R^{3 \times 3} \\
\phi_k(\mathbf p_m) &= \psi(\mathbf p_m) \in \mathbb R^{3 \times 3} \\
\end{split}
\end{equation}
A related approach for SE(3) and SO(2) is described in \citet{gta_miyato_2024}.
In contrast to RoPE, using the $x$ and $y$ coordinates directly causes training instability if the position magnitudes are large, which can be mitigated by downscaling the positions \cite{scene_diffuser_jiang_2024}.
Even with this downscaling, we find that a sine-cosine encoding for $x$ and $y$ yields better performance.

\section{Method} \label{sec:method}

\subsection{Objective}
Based on the strong performance of RoPE \cite{roformer_su_2024}, we hypothesize that a block-diagonal matrix of 2D rotations would perform well for SE(2) invariant attention:
\begin{equation}
\phi(\mathbf p_{n \to m}) = \textrm{diag} \left[\rho (x_{n \to m}), \rho (y_{n \to m}), \rho(\theta_{n \to m}) \right] 
\end{equation}
Note that the relative angle $\theta_{n \to m} = \theta_m - \theta_n$, so we can use standard RoPE for that block.
The relative x and y blocks are more challenging, because the equations for $x_{n \to m}$ and $y_{n \to m}$ involve $x_n, y_n, \theta_n, x_m$ and $y_m$.

\subsection{Fourier Approximation}
First consider the relative x block $\rho (x_{n \to m})$.
The full expression for the relative x position is $x_{n \to m} = (x_m - x_n) \cos \theta_n + (y_m - y_n) \sin \theta_n$.
To follow RoPE, we need to separate this expression into a sum of two terms, where the first term depends only on $\mathbf p_n$ and the second term depends only on $\mathbf p_m$.
This is clearly not possible; terms such as $x_m \cos \theta_n$ involve both query and key poses.
However, we can approximately factorize $\rho(x_{n \to m})$ using a Fourier series.
We start by separating $x_{n \to m}$ into two terms, the first depending on $\mathbf p_n$ and the second depending on $\theta_n$, $x_m$, and $y_m$:
\begin{equation} \label{eqn:split_x_rel}
\begin{split}
&x_{n \to m} = \\
&\; \underbrace{\left( -x_n \cos \theta_n - y_n \sin \theta_n\right)}_{v^{(x)}_n}+ \underbrace{\left(x_m \cos \theta_n + y_m \sin \theta_n\right)}_{u^{(x)}_{m}(\theta_n)}
\end{split}
\end{equation}
Therefore, $\rho \left(x_{n \to m} \right) = \rho \left( v_n^{(x)} \right) \rho \left( u_{m}^{(x)}(\theta_n) \right)$.
While we can't factorize the second term exactly, we can approximate it using a Fourier series. 
Consider the functions $\{g_i \}_{i \in \mathbb N}$ where
\begin{equation}
g_i (z) = \begin{cases}
\cos \left(\frac{i}{2} z\right) & i \textrm{ even} \\
\sin \left(\frac{i+1}{2} z\right) & i \textrm{ odd}
\end{cases}
\end{equation}
We can approximate 
\begin{equation}
\cos \left( u_{m}^{(x)} (\theta_n) \right) \approx \sum_{i=0}^{F-1} \Gamma^{(x)}_m (i) \cdot g_i(\theta_n)
\end{equation}
where $F \in \mathbb N$ is the number of terms in the approximation and the coefficients $\Gamma^{(x)}_m$ are a function of $x_m$ and $y_m$:
\begin{equation}
\Gamma^{(x)}_m (i) = \frac{a_i}{2 \pi} \int_{-\pi}^\pi \cos \left( u_m^{(x)} (z) \right) g_i (z) dz
\end{equation}
where $a_i = 1$ if $i = 0$ and 2 otherwise.
This coefficient can be computed using numerical integration with $2F$ points.
We can perform a similar approximation for the sine term, where the only change is the coefficients:
\begin{equation}
\Lambda^{(x)}_m (i) = \frac{a_i}{2 \pi} \int_{-\pi}^\pi \sin \left( u_m^{(x)} (z) \right) g_i (z) dz
\end{equation}
Using $\mathbf {\Gamma}^{(x)}_m$ and $\mathbf \Lambda^{(x)}_m$ as the vectors of coefficients and $\mathbf b_n = \begin{bmatrix} g_0(\theta_n) & g_1(\theta_n) & \ldots & g_{F-1} (\theta_n) \end{bmatrix}$, we can approximate
\begin{equation}
\rho \left( u^{(x)}_m (\theta_n) \right) \approx \underbrace{ \begin{bmatrix} \mathbf b_n^\top & \mathbf 0 \\ \mathbf 0 & \mathbf b_n^\top \end{bmatrix} }_{\mathbb R^{2 \times 2F}} \underbrace{ \begin{bmatrix} \mathbf \Gamma^{(x)}_m & - \mathbf \Lambda_m^{(x)} \\ \mathbf \Lambda_m^{(x)} & \mathbf \Gamma^{(x)}_m \end{bmatrix} }_{\mathbb R^{2F \times 2}}
\end{equation}
Combining this with Equation \ref{eqn:split_x_rel}, we get that
\begin{equation}
\begin{split}
&\rho (x_{n \to m}) \approx \\
&\; \begin{bmatrix} \cos \left( v^{(x)}_n \right) \mathbf b_n^\top & - \sin \left( v^{(x)}_n \right) \mathbf b_n^\top \\ \sin \left( v^{(x)}_n \right) \mathbf b_n^\top & \cos \left( v^{(x)}_n \right) \mathbf b_n^\top \end{bmatrix} \begin{bmatrix} \mathbf \Gamma^{(x)}_m & - \mathbf \Lambda_m^{(x)} \\ \mathbf \Lambda_m^{(x)} & \mathbf \Gamma^{(x)}_m \end{bmatrix}
\end{split}
\end{equation}

A similar approach can be used for $\rho (y_{n \to m})$ using 
\begin{equation}
\begin{split}
&y_{n \to m} = \\
&\; \underbrace{\left( x_n \sin \theta_n - y_n \cos \theta_n \right)}_{v^{(y)}_n}+ \underbrace{\left(-x_m \sin \theta_n + y_m \cos \theta_n\right)}_{u^{(y)}_{m}(\theta_n)}
\end{split}
\end{equation}

\subsection{SE(2) Fourier}
We propose the following implementation of Algorithm \ref{alg:relative-attention} for $\mathcal G = \textrm{SE}(2)$:
\begin{equation}
\begin{split}
\phi_q (\mathbf p_n) &= \textrm{diag} \left[ \phi_q^{(x)} (\mathbf p_n), \phi_q^{(y)} (\mathbf p_n), \phi_q^{(\theta)} (\mathbf p_n) \right] \\
\phi_q^{(x)} (\mathbf p_n) &= \begin{bmatrix} \cos \left( v^{(x)}_n \right) \mathbf b_n^\top & - \sin \left( v^{(x)}_n \right) \mathbf b_n^\top \\ \sin \left( v^{(x)}_n \right) \mathbf b_n^\top & \cos \left( v^{(x)}_n \right) \mathbf b_n^\top \end{bmatrix} \\
\phi_q^{(y)} (\mathbf p_n) &= \begin{bmatrix} \cos \left( v^{(y)}_n \right) \mathbf b_n^\top & - \sin \left( v^{(y)}_n \right) \mathbf b_n^\top \\ \sin \left( v^{(y)}_n \right) \mathbf b_n^\top & \cos \left( v^{(y)}_n \right) \mathbf b_n^\top \end{bmatrix} \\
\phi_q^{(\theta)} (\mathbf p_n) &= \rho (-\theta_n) \\
\phi_k (\mathbf p_m) &= \textrm{diag} \left[ \phi_k^{(x)} (\mathbf p_m), \phi_k^{(y)} (\mathbf p_m), \phi_k^{(\theta)} (\mathbf p_m) \right] \\
\phi_k^{(x)} (\mathbf p_m) &= \begin{bmatrix} \mathbf \Gamma^{(x)}_m & - \mathbf \Lambda_m^{(x)} \\ \mathbf \Lambda_m^{(x)} & \mathbf \Gamma^{(x)}_m \end{bmatrix} \\
\phi_k^{(y)} (\mathbf p_m) &= \begin{bmatrix} \mathbf \Gamma^{(y)}_m & - \mathbf \Lambda_m^{(y)} \\ \mathbf \Lambda_m^{(y)} & \mathbf \Gamma^{(y)}_m \end{bmatrix} \\
\phi_k^{(\theta)} (\mathbf p_m) &= \rho (\theta_m). \\
\end{split}
\end{equation}
In contrast to RoPE, these matrices are no longer square: $\phi_q (\mathbf p_n) \in \mathbb R^{6 \times (4F + 2)}$ and $\phi_k (\mathbf p_m) \in \mathbb R^{(4F + 2) \times 6}$.
With these matrices,
\begin{equation}
\begin{split}
\phi_q &(\mathbf p_n) \phi_k (\mathbf p_m) \approx \\
&\textrm{diag} \left[ \rho (x_{n \to m}), \rho (y_{n \to m}), \rho (\theta_{n \to m}) \right],
\end{split}
\end{equation}
we can scale the x and y components before computing these matrices to obtain features at different spatial resolutions. \cite{fourier_features_tancik_2020}

\section{Experiments}

\begin{figure}[ht]
    \centering
    \includegraphics[width=0.48\textwidth]{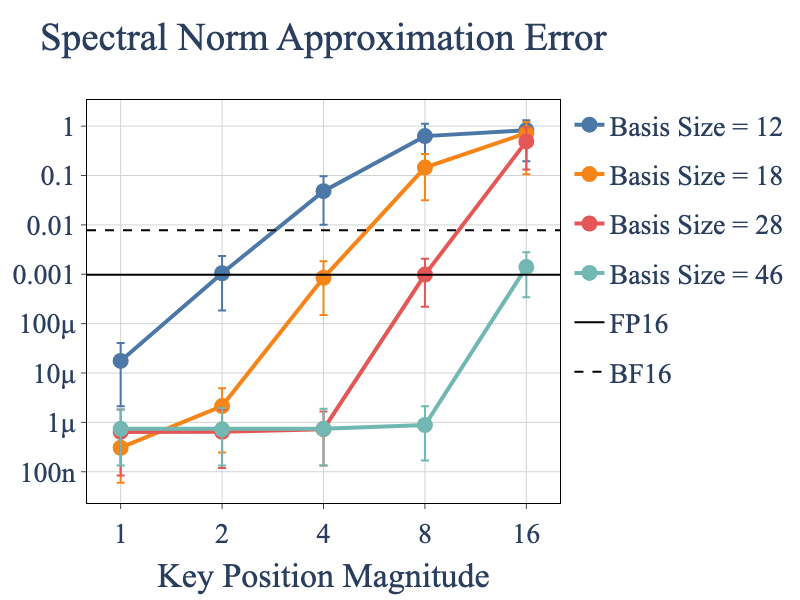}
    \caption{Spectral norm of the approximation error $\lVert \phi(\mathbf p_{n \to m}) - \phi_q (\mathbf p_n) \phi_k (\mathbf p_m) \rVert_2$ computed with 32-bit floating point. Key positions are sampled uniformly from the circle with radius described by the x-axis, and query headings are sampled uniformly from $[0, 2\pi]$. Lines denote the average error with error bars to denote the 2.5\% and 97.5\% percentiles. The horizontal lines for FP16 and BF16 denote the precision of the quadratic representation using 16-bit floating point and bfloat-16 for values with absolute value 1 (i.e. the smallest $\epsilon > 0$ such that $1+\epsilon$ can be expressed in that floating point format). If the position magnitudes are $\le 4$, we can achieve an approximation error comparable to the inherent precision of 16-bit floating point arithmetic with only 18 terms. As shown by the chosen basis sizes, to maintain a spectral norm approximation error around $10^{-3}$ the basis size needs to increase by roughly 50\% when doubling the magnitude of key positions.}
    \label{fig:spectral-norm-approx-error}
\end{figure}

We measure the accuracy of our SE(2)-invariant approach using the Fourier approximation error $\phi(\mathbf p_{n \to m}) - \phi_q (\mathbf p_n) \phi_k (\mathbf p_m)$ using the spectral norm on an abstract toy problem. 

\subsection{Approximation Error}
The term being approximated depends on the key position $(x_m, y_m)$, the query heading $\theta_n$, and the number of basis elements $F$.
In Figure \ref{fig:spectral-norm-approx-error} we measure the approximation error between the desired matrix $\phi(\mathbf p_{n \to m})$ and the approximate matrix $\phi_q(\mathbf p_n) \phi_q (\mathbf k_m)$ by sampling $(x_m, y_m)$ uniformly from a circle of a fixed radius and $\theta_n$ uniformly from $[0, 2\pi]$.

\begin{figure}
    \centering
    \includegraphics[width=0.5\textwidth]{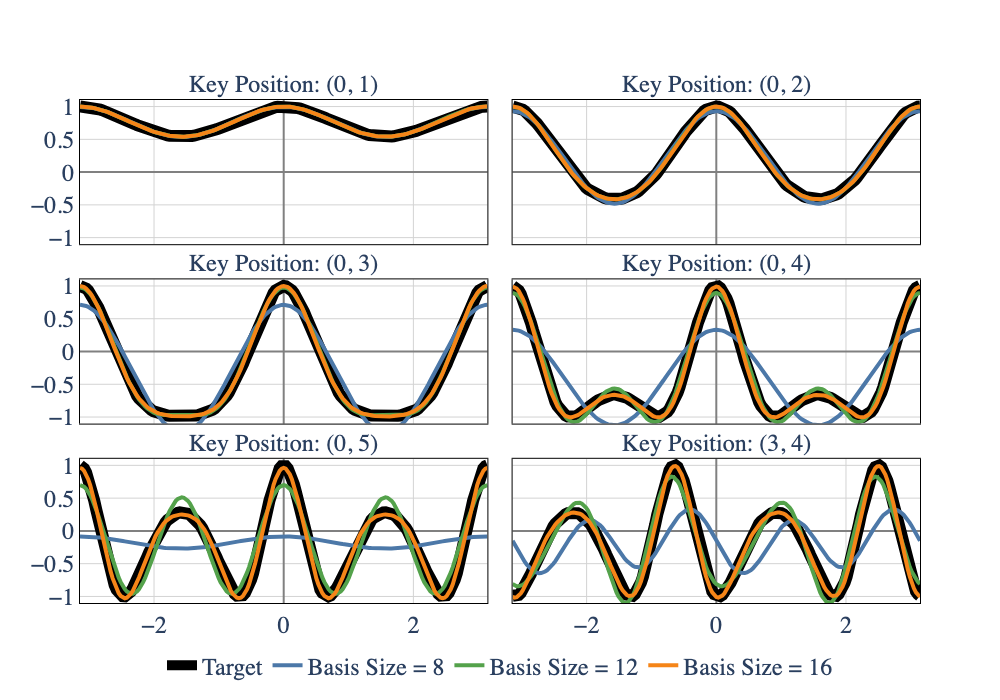}
    \caption{Plots of the target function $\cos \left( u_m^{(x)} (\theta) \right)$ for different key positions $(x_m, y_m)$, along with their Fourier series approximations for various basis sizes. As the key position magnitude increases, the target function gains higher frequency components and requires additional series terms to approximate accurately. Rotating the key position about the origin while keeping the magnitude constant shifts the target function, but the Fourier approximations can change in other ways.}
    \label{fig:function-plots}
\end{figure}

For any fixed basis size, the approximation error gets worse as the key position magnitude increases.
Larger key position magnitudes cause the target functions such as $\cos \left( u_m^{(x)}(\theta) \right)$ to have higher frequency components, requiring more terms from the Fourier series, as shown in Figure \ref{fig:function-plots}.
The same pattern holds for $u_m^{(y)}$ instead of $u_m^{(x)}$ and for $\sin$ instead of $\cos$.

Using a basis size of 12, 18, and 28 achieves an average approximation error comparable to the quadratic representation with 16-bit floating point precision when the key position magnitude is 2, 4, and 8, respectively.
Flash Attention \cite{flashattention_dao_2022} is generally implemented for float-16 \cite{mixed_precision_micikevicius_2018} and bfloat-16 \cite{bfloat16_Wang_Kanwar_2019}, so this level of approximation error does not significantly change the overall numerical accuracy of the model.
Furthermore, existing Flash Attention implementations support feature dimensions of at least 256, enough to accommodate several blocks of SE(2) Fourier with these basis sizes.

\subsection{Analysis on Agent Simulation}

We additionally evaluated our representation on mechanisms for agent simulation using a private dataset containing 33M scenarios. The agent simulation task \cite{waymo_sim_agents_montali_2023} requires predicting the next action for a set of agents conditioned on the agents' history, road map, and traffic signals.
We use a next token prediction model \cite{smart_wu_2024, closed_loop_fine_tuning_zhang_2025, behavior_gpt_zhou_2024} where the agents (e.g. vehicles, pedestrians) and map elements (e.g. driving lanes, crosswalks) are tokenized and processed with a transformer \cite{attention_all_you_need_vaswani_2017} to then predict a categorical distribution over a discrete set of actions per agent.
Each token corresponds to a specific agent or map element and has an associated SE(2) pose.
Positions are downscaled to have magnitude $\le 4$.
In our experiments we use the same model architecture with the only change being the relative attention mechanism.
Details on the model architecture are provided in the supplemental material. In addition to the three forms of relative attention described above, we also compare against absolute position embeddings.
This approach adds an embedding of the token pose to the token feature vector and then uses standard SDPA.

Negative log likelihood (NLL) measures the likelihood of the ground truth actions predicted by the model.
Minimum average displacement error (minADE) samples 16 6-second joint trajectories.
For each sample, the average distance to the ground truth trajectory is measured (average displacement error), and we then take the minimum across all samples.
Because the magnitude of prediction error depends on the type of trajectory, we divide minADE based on the ground truth future trajectory into three categories: stationary, straight, and turning.

Table \ref{tab:traj-metrics} shows that 2D RoPE and SE(2) Fourier outperform both absolute positions and the SE(2) Representation approach.
While 2D RoPE slightly outperforms SE(2) Fourier on straight trajectories, SE(2) Fourier significantly outperforms 2D RoPE on turning trajectories, the hardest category.

\begin{table}[]
    \centering
    \begin{tabular}{ccccc}
    \toprule
      \shortstack{Attention \\ Method} & NLL & \shortstack{Stationary \\ minADE} & \shortstack{Straight \\ minADE} & \shortstack{Turning \\ minADE} \\
        \midrule
      Absolute Positions & 0.193 & 0.24 & 1.90 & 2.98 \\
      2D RoPE & \textbf{0.190} & \textbf{0.23} & \textbf{1.78} & 2.69 \\
      SE(2) Representation & 0.191 & \textbf{0.23} & 1.82 & 2.70 \\
      SE(2) Fourier \textit{(ours)} & \textbf{0.190} & \textbf{0.23} & 1.79 & \textbf{2.60} \\
    \bottomrule
    \end{tabular}
    \caption{Metrics for agent simulation. Reported numbers are the mean across 3 experiments with different random initialization.}
    \label{tab:traj-metrics}
\end{table}

\section{Limitations \& Conclusion} 
In this work we present SE(2) Fourier, a novel approach for SE(2)-invariant scaled dot-product attention using linear GPU memory.
We show the mathematical derivation, evaluate the accuracy of the approximation relative to a baseline  representation that is quadratic in memory, and demonstrate its benefit for an agent simulation task.
In future work, we will demonstrate the benefit of this approach on a wider range of driving tasks and datasets.
We will also include more ablation experiments comparing our method against other approaches, such as data augmentation and quadratic memory SE(2) invariant attention).
While we focus on autonomous driving, our hope is that this approach will benefit a wide range of robotics tasks where SE(2) invariance is desirable, such as ground-based navigation.
However, there are also many robotics tasks where SE(3) poses are used.
It is an open question whether a similar strategy can be used for linear memory SE(3) invariant attention.

\bibliographystyle{plainnat}
\bibliography{main}

\end{document}